\documentclass{article}
\usepackage[utf8]{inputenc}
\usepackage{spconf, amsmath, graphicx, hyperref}
\usepackage{amsfonts, amssymb}
\usepackage{booktabs}
\usepackage{algorithm}
\usepackage{algpseudocode}
\usepackage{lipsum}
\usepackage{lineno}
\usepackage{float}
\usepackage{wrapfig}
\usepackage{pifont}
\usepackage{tikz}
\definecolor{darkblue}{rgb}{0,  0,  0.5}
\definecolor{DarkGreen}{RGB}{34,  139,  34}
\definecolor{DarkPurple}{RGB}{142,  124,  195}
\definecolor{DarkOrange}{RGB}{246,  178,  107}
\definecolor{LightGreen}{RGB}{80,  170,  80}
\newcommand{\Yes}{\textcolor{LightGreen}{\ding{51}}} % Green Heart
\newcommand{\No}{\textcolor{red}{\ding{55}}}

\makeatletter
\renewcommand{\@makefnmark}{\hbox{\@textsuperscript{\fnsymbol{footnote}}}}
\makeatother

\title{TsqLoRA: Towards Sensitivity and Quality Low-Rank Adaptation for Efficient Fine-Tuning}
\name{Yu Chen$^{1}$,  Yifei Han$^2$,  Long Zhang$^2$,  Yue Du$^2$,  Bin Li$^{2\textasteriskcentered}$}

\vspace{-3em}

\address{$^1$South China University of Technology,  \\ $^2$Shenzhen Institute of Advanced Technology,  Chinese Academy of Sciences}
\begin{document}
%\ninept
%
\maketitle
\begin{abstract}
Fine-tuning large pre-trained models for downstream tasks has become a fundamental approach in natural language processing. Fully fine-tuning all model parameters is computationally expensive and memory-intensive,  especially in resource-constrained environments. Existing parameter-efficient fine-tuning methods reduce the number of trainable parameters but typically overlook the varying sensitivity of different model layers and the importance of training data. In this work,  we propose \textbf{TsqLoRA},  a novel method that integrates data-quality-driven selection with sensitivity-aware low-rank adaptation,  consisted of two main components: a quality-aware sampling mechanism for selecting the most informative training data,  and a dynamic rank allocation module that adjusts the rank of each layer based on its sensitivity to parameter updates. The experimental results demonstrate that TsqLoRA improves fine-tuning efficiency while maintaining or even improving performance on a variety of NLP tasks. Our code will be available at \url{https://github.com/Benjamin-Ricky/TsqLoRA}.
\end{abstract}
\vspace{-0.2cm}
\vspace{-0.3cm}
\section{Introduction}
\label{sec:intro}
\vspace{-0.1cm}
\footnotetext{$^{*}$Corresponding author.}
Large pre-trained language models (PLMs) such as BERT~\cite{devlin2019bert},  GPT~\cite{Radford2018ImprovingLU} have demonstrated remarkable success in a wide range of natural language processing (NLP) tasks,  from natural language understanding benchmarks like GLUE~\cite{wang2018glue} to natural language generation datasets such as XSum~\cite{narayan2018don}. Fine-tuning these large-scale models for downstream tasks has thus become a central paradigm. However,  fully fine-tuning all model parameters is computationally expensive and memory-intensive,  which significantly limits deployment in resource-constrained environments. Therefore,  developing parameter-efficient fine-tuning (PEFT) methods has become a critical research direction~\cite{houlsby2019parameter,  ding2023parameter}. 

\begin{figure}[t]
  \centering
  \includegraphics[width=8.5cm]{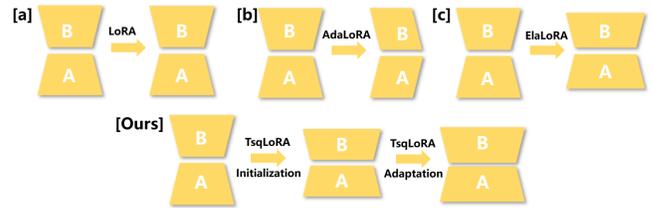}
\vspace{-0.3cm}
  \caption{Comparison of TsqLoRA,  LoRA~\cite{hu2022lora},  AdaLoRA~\cite{Zhang2023AdaLoRA} and ElaLoRA~\cite{chang2025elalora}. LoRA performs fine-tuning with fixed ranks,  AdaLoRA prunes ranks,  ElaLoRA prunes ranks while expanding them. While TsqLoRA initializes ranks with data quality and adjusts them during training.}
  \vspace{-0.5cm}
  \label{fig:comparison_overview}
\end{figure}

Despite significant progress,  several challenges remain. Existing PEFT methods typically focus on designing efficient adaptation modules but often ignore two crucial aspects: (1) the varying \emph{sensitivity} of different model layers to parameter updates,  and (2) the \emph{quality and informativeness of the training data}. For example,  LoRA~\cite{hu2022lora} (as shown in Fig~\ref{fig:comparison_overview}[a]) introduces low-rank adapters to reduce the number of trainable parameters,  while AdaLoRA~\cite{Zhang2023AdaLoRA}(Fig~\ref{fig:comparison_overview}[b]) adaptively adjusts rank budgets based on importance scores,  and ElaLoRA~\cite{chang2025elalora}(Fig~\ref{fig:comparison_overview}[c]) both prunes and expands ranks. However,  these methods primarily rely on parameter-side optimization and assume the entire training dataset is equally useful,  which may lead to inefficiency when noisy or redundant samples dominate. 

Similarly,  data selection methods such as TS-DSHA-PLEY~\cite{Schoch2023} and 0.5\% Data ~\cite{Chen2023small} demonstrate that carefully chosen subsets can match or surpass full-dataset fine-tuning. Yet,  these approaches are typically decoupled from PEFT frameworks and do not consider how data quality interacts with sensitivity-aware parameter adaptation.

To solve the above problem,  we propose \textbf{TsqLoRA},  as shown in Fig~\ref{fig:comparison_overview}
[Ours],  a data-quality-driven and sensitivity-aware low-rank adaptation method that unifies data selection and parameter-efficient fine-tuning. Our method introduces two central innovations. First,  it incorporates a quality-aware sampling strategy that prioritizes informative data and reduces redundancy in training. Second,  it employs a sensitivity-guided dynamic rank allocation scheme that flexibly adjusts rank capacity according to the importance of different layers,  ensuring efficient use of limited parameters. Together,  these two mechanisms make TsqLoRA substantially more effective than prior approaches. Extensive experiments on GLUE and XSum benchmarks further demonstrate that TsqLoRA achieves state-of-the-art results among parameter-efficient fine-tuning methods,  especially in low-rank and low-data settings,  significantly reducing computational cost while maintaining or improving accuracy.

The contributions of this paper can be summarized as follows:
\begin{itemize}
    \item A quality-aware sampling strategy that prioritizes informative data and reduces redundancy.
    \item A sensitivity-guided dynamic rank allocation scheme that adapts to the importance of the layer.
    \item We conduct extensive experiments across multiple benchmarks under different parameter budgets,  demonstrating the effectiveness of TsqLoRA.
\end{itemize}
\vspace{-0.6cm}
\section{Methodology}
\label{sec:methodology}
\vspace{-0.2cm}
We propose TsqLoRA,  a data-quality-driven and sensitivity-aware low-rank adaptation method. The key idea is to jointly model which data to use and how to allocate adaptation capacity,  The full ElaLoRA training procedure is outlined in Algorithm ~\ref{alg:TsqLoRA1}. 
\vspace{-0.5cm}
\subsection{Data Quality Estimation}
\vspace{-0.2cm}
We define the quality of a sample $(x_i, y_i)$ as its contribution to model improvement,  measured through gradient norms,  loss reduction,  and convergence acceleration. Specifically, 
\begin{equation}
Q(x_i) = \alpha_1 \|\nabla_\theta \ell(x_i, y_i;\theta)\|_2 + \alpha_2 \Delta \ell(x_i) + \alpha_3 \Delta C(x_i), 
\end{equation}
where $\Delta \ell(x_i) = \ell(\theta) - \ell(\theta - \eta \nabla_\theta \ell(x_i, y_i;\theta))$ is the potential loss reduction,  and $\Delta C(x_i)$ denotes the contribution to convergence speed. The combined quality score integrates multiple perspectives without relying on heuristics such as lexical diversity.

Data selection is performed by weighted sampling:
\begin{equation}
p(x_i) = \frac{\exp(Q(x_i)/\tau)}{\sum_j \exp(Q(x_j)/\tau)}, 
\end{equation}
where $\tau$ is a temperature parameter controlling the sharpness of the distribution.
\vspace{-0.5cm}
\subsection{Layer Sensitivity Analysis}
\vspace{-0.2cm}
We compute layer sensitivity as the importance of its parameters under the current training distribution. For layer $\ell$ with weight $W_\ell$,  the sensitivity score is defined as
\begin{equation}
S_\ell = \mathbb{E}_{(x, y)\sim\mathcal{S}} \Big[ | W_\ell \odot \nabla_{W_\ell} \ell(x, y;\theta) | \Big], 
\end{equation}
where $\odot$ denotes element-wise multiplication. This definition corresponds to the expected gradient-weight product,  highlighting parameters with both high magnitude and strong update signals.

Alternatively,  sensitivity can be estimated by the Fisher information:
\begin{equation}
F_\ell = \mathbb{E}_{(x, y)\sim\mathcal{S}} \big[ \nabla_{W_\ell} \ell(x, y;\theta)^\top \nabla_{W_\ell} \ell(x, y;\theta) \big], 
\end{equation}
and we normalize $S_\ell \in (0, 1)$ to enable stable comparisons across layers.

\vspace{-0.5cm}
\subsection{Dynamic Rank Allocation}
\vspace{-0.2cm}
Each LoRA update is parameterized as
\begin{equation}
\Delta W_\ell = B_\ell A_\ell,  \qquad A_\ell \in \mathbb{R}^{d\times r_\ell},  \; B_\ell \in \mathbb{R}^{r_\ell \times d}, 
\end{equation}
where $r_\ell$ is the rank for layer $\ell$. Instead of fixing $r_\ell$,  TsqLoRA dynamically adjusts it according to both data quality and sensitivity:
\begin{equation}
r_\ell^{(t)} = \frac{S_\ell}{\sum_j S_j} \cdot \phi(Q(\mathcal{S})) \cdot R_0, 
\end{equation}
where $R_0$ is a base rank budget and $\phi(Q(\mathcal{S}))$ is a scaling factor proportional to the average data quality in the selected set $\mathcal{S}$.

\begin{algorithm}[t]
\caption{TsqLoRA}  % 新增标题
\label{alg:TsqLoRA1}
\begin{algorithmic}[1]
\Require Pre-trained model $f_{\theta_0}$; training pool $\mathcal{D}$; iterations $I$; adjustment schedule $T$.
\State \textbf{Initialize} LoRA modules with base rank $r_0$; set moving stats for importance/sensitivity.
\State \textbf{Warm-up data scoring:} run a short pass on $\mathcal{D}$ to obtain per-sample quality scores $Q(x)$; construct weighted subset $\mathcal{S}$.
\For{$i=1$ to $I$}
\State Sample minibatch $\mathcal{B} \subset \mathcal{S}$ by probabilities proportional to $Q(\cdot)$; compute loss and backprop.
\State Update LoRA parameters on each adapted layer; update statistics of layer sensitivity/importance (e.g.,  grad–weight products).
\If{$i \in T$}
\State Re-compute layer sensitivities and normalize to $(0, 1)$.
\State Allocate new ranks for layers based on sensitivity (and current average data quality); apply mask/expand to LoRA factors.
\State Optionally refresh $Q(x)$ on a light pass and reweight $\mathcal{S}$.
\EndIf
\EndFor
\State \Return fine-tuned model $f_{\theta}$.

\end{algorithmic}
\end{algorithm}

\vspace{-0.4cm}
\section{Experiments and Results}
\label{sec:expandres}
\vspace{-0.2cm}
\subsection{Experimental Setup}
% \vspace{-0.2cm}
We evaluate our proposed TsqLoRA method using the DeBER
Ta-v3-base~\cite{He2021DeBERTaV3ID} model as the backbone. Experiments are primarily conducted on the GLUE~\cite{wang2018glue} benchmark,  which consists of multiple natural language understanding tasks such as sentiment classification,  paraphrase detection,  and textual entailment. We analyze results from these three perspectives: (1) comparison with strong LoRA-family baselines including LoRA \cite{hu2022lora},  AdaLoRA \cite{Zhang2023AdaLoRA},  and ElaLoRA \cite{chang2025elalora}; (2) comparison against representative data selection methods such as TS-DSHAPLEY \cite{Schoch2023},  0.5\% Data \cite{Chen2023small},  and LESS \cite{Xia2024LESS}; and (3) ablation of key components in TsqLoRA. All experiments were conducted on NVIDIA A100 GPUs with 80GB memory,  using the PyTorch framework \cite{Paszke2019PyTorch}. Implementation leveraged the Hugging Face Transformers library for preprocessing and model initialization \cite{wolf2019huggingface}.
\vspace{-0.6cm}
\subsection{Natural Language Understanding Task}
\begin{figure}[th!]
  \centering
  \includegraphics[width=8.5cm]{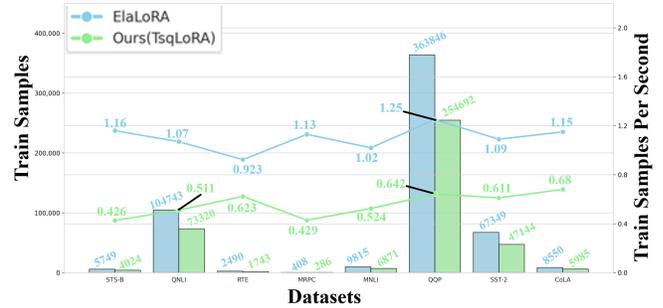}
  \vspace{-0.4cm}
  \caption{Comparison of features of training with the number of train samples and train samples per second across GLUE~\cite{wang2018glue} benchmark with ElaLoRA~\cite{chang2025elalora},  reporting the average results.}
  \label{fig:scaled_comparison}
\end{figure}

\begin{figure}[t!]
  \centering
  \includegraphics[width=\linewidth]{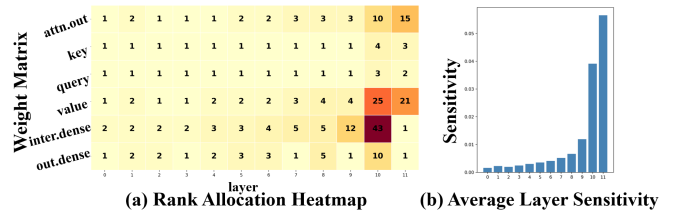}
  \vspace{-0.6cm}
   \caption{QNLI: Final Rank Distribution vs Layer Sensitivity,  where ranks are initialized by 10.}
   \label{fig:rank_heatmaps}
   \vspace{-0.2cm}
\end{figure}

\begin{figure}[t!]
  \centering
  \includegraphics[width=8.5cm]{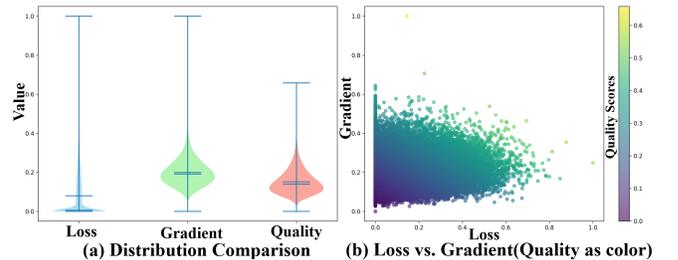}
  \vspace{-0.4cm}
  \caption{Features of distributions of loss contribution,  gradient contribution and quality scores during QNLI task.}
  \label{fig:quality_scores}
  \vspace{-0.5cm}
\end{figure}

\vspace{-0.2cm}
\noindent \textbf{Main Results.}
We report performance on the development set of GLUE~\cite{wang2018glue}. For MNLI,  QNLI,  SST-2,  RTE,  CoLA,  and MRPC,  we report accuracy; for QQP and MRPC we also include F1; and for STS-B we report Pearson correlation. Table~\ref{tab:glue_ranks} summarizes the results on the GLUE benchmark. TsqLoRA consistently achieves the highest average performance across the three tested rank settings. In particular,  TsqLoRA with rank $r=2$ already outperforms both AdaLoRA \cite{Zhang2023AdaLoRA} and ElaLoRA \cite{chang2025elalora}. This demonstrates that incorporating data quality awareness and sensitivity-guided adaptation leads to more effective utilization of limited parameters. When the rank increases to $r=4$,  TsqLoRA maintains superior average accuracy,  showing stable improvements across SST-2,  QQP,  and RTE. At $r=10$,  TsqLoRA achieves the best overall average,  slightly surpassing ElaLoRA,  while attaining competitive or superior performance on most individual tasks. Not only our method outperforms others across GLUE benchmark,  as shown in Fig.~\ref{fig:scaled_comparison},  compared with ElaLoRA,  our method utilizes fewer train samples and takes much less time to train. We also plot TsqLoRA's final rank distribution (as shown in Fig.~\ref{fig:rank_heatmaps}),  showing that layers with higher sensitivity are allocated higher ranks. Additionally,  we further analyze the features of distributions of loss contribution,  gradient contribution and quality scores by picking QNLI task as an example,  as you can see in Fig.~\ref{fig:quality_scores},  quality scores are shown in a gaussian-like distribution,  which is the same as loss distribution and gradient contribution,  highlighting that while quality scores align with loss and gradient distributions,  their specific impact requires careful integration.

\begin{table}
  \small
  \centering
  \begin{tabular}{lccc}
    \toprule
    Batch Size  & Epochs & $\tau$ \\
    \midrule
    64 & 50 & 0.7 \\
    \bottomrule
  \end{tabular}
  \caption{Hyperparameters used for GLUE tasks.}
  \vspace{-0.6cm}
  \label{tab:glue_config}
\end{table}

\noindent \textbf{Comparison with Data Selection Methods.}
Table~\ref{tab:glue_simplified} compares TsqLoRA with representative data selection methods on SST-2,  QQP,  and RTE. TsqLoRA achieves 95.6 on SST-2,  92.0 on QQP,  and 86.5 on RTE,  outperforming prior data selection baselines by a significant margin. This demonstrates that TsqLoRA not only reduces parameter inefficiency but also improves robustness under reduced or filtered datasets.

% ===================== ICASSP-style triple-line table =====================
\begin{table*}
  \small
  \centering
  \begin{tabular*}{\linewidth}{lccccccccccc}
    \toprule
    \textbf{Method} & \textbf{\# Params} & \textbf{Rank} & \textbf{MNLI} & \textbf{SST-2} & \textbf{CoLA} & \textbf{QQP} & \textbf{QNLI} & \textbf{RTE} & \textbf{MRPC} & \textbf{STS-B} & \textbf{All} \\
        & & & \textbf{Acc.} & \textbf{Acc.} & \textbf{Mcc} & \textbf{Acc./F1} & \textbf{Acc.} & \textbf{Acc.} & \textbf{Acc.} & \textbf{Corr.} & \textbf{Avg.} \\
    \midrule
    LoRA & 0.33M & & 88.43 & 94.49 & \textbf{69.77} & 90.92/88.02 & \textbf{93.84} & 83.03 & 88.48 & 91.13 & 87.51 \\
    AdaLoRA & 0.49M & & 89.05 & 94.95 & 66.87 & 90.99/88.11 & 94.33 & 84.84 & 86.51 & \textbf{91.60} & 87.39 \\
    ElaLoRA & 0.33M & & \textbf{89.32} & \textbf{95.53} & 67.38 & 91.21/88.33 & 93.70 & 85.92 & \textbf{90.44} & 90.61 & 88.01 \\
    \textbf{Ours(TsqLoRA)} & 0.33M & r = 2 & \textbf{89.32} & 95.07 & 69.51 & \textbf{91.40/88.42} & 93.57 & \textbf{86.28} & 88.24 & 90.67 & \textbf{88.09} \\
    \midrule
    LoRA & 0.66M &  & 89.46 & 94.15 & 67.15 & 91.26/88.50 & 93.84 & 85.55 & 88.48 & 91.17 & 87.63 \\
    AdaLoRA & 0.99M &  & 88.92 & 95.53 & 67.70 & 91.10/88.25 & 94.12 & 86.28 & 87.74 & \textbf{91.74} & 87.89 \\
    ElaLoRA & 0.66M &  & 89.44 & \textbf{95.64} & \textbf{70.15} & 91.26/88.60 & 94.44 & 87.36 & 90.44 & 91.10 & 88.72 \\
    \textbf{Ours(TsqLoRA)} & 0.66M & r = 4 & \textbf{89.46} & 95.30 & 70.00 & \textbf{91.45/88.81} & \textbf{94.51} & \textbf{87.36} & \textbf{90.69} & 90.93 & \textbf{88.72} \\
    \midrule
    LoRA & 1.66M & & 89.11 & 92.31 & 67.06 & 91.34/88.62 & 94.08 & 87.72 & 88.97 & 91.40 & 87.75 \\
    AdaLoRA & 2.49M &  & 89.27 & 95.76 & 70.02 & 91.16/88.37 & 94.31 & 87.73 & 88.72 & \textbf{91.67} & 88.58 \\
    ElaLoRA & 1.66M &  & \textbf{89.51} & 95.87 & 70.19 & 91.42/88.83 & \textbf{94.36} & \textbf{88.81} & 88.98 & 91.59 & 88.84 \\
    \textbf{Ours(TsqLoRA)} & 1.66M & r = 10 & 89.47 & \textbf{96.10} & \textbf{70.28} & \textbf{91.47/88.88} & 94.34 & 88.45 & \textbf{89.46} & 91.33 & \textbf{88.86} \\
    \bottomrule
  \end{tabular*}
\vspace{-0.2cm}
   \caption{GLUE dev results with DeBERTa-v3-base~\cite{He2021DeBERTaV3ID} under different LoRA ranks. QQP and MRPC report Accuracy/F1; STS-B reports Pearson Correlation(\%). Baseline results are from ElaLoRA,  ours are reproduced with our latest experiments.}

\vspace{-0.2cm}
   \label{tab:glue_ranks}
\end{table*}

% ========================================================================

\begin{table}
  \small
  \resizebox{\linewidth}{!}{
  \begin{tabular}{lccc}
    \toprule
    \textbf{Method} & \textbf{SST-2 (Acc.)} & \textbf{QQP (Acc.)} & \textbf{RTE (Acc.)} \\
    \midrule
    TS-DSHAPLEY \cite{Schoch2023} & 95.30 & 91.90 & 80.10 \\
    0.5\% Data \cite{Chen2023small} & 94.39 & 84.66 & 74.70 \\
    AlpaGasus \cite{Chen2023Alp} & 95.90 & 91.95 & 86.34 \\
    LESS \cite{Xia2024LESS} & 94.81 & 85.42 & 79.56 \\
    LIMA \cite{Zhou2023LIMA} & 95.30 & 90.92 & 85.42 \\
    Instruction Mining \cite{Cao2024InstrMining} & 96.00 & 91.02 & 84.56 \\
    \midrule
   \textbf{Ours(TsqLoRA)} & \textbf{96.00} & \textbf{92.00} & \textbf{86.50} \\
    \bottomrule
  \end{tabular}
  }
  \vspace{-0.2cm}
  \caption{Simplified comparison of GLUE~\cite{wang2018glue} dev results (SST-2,  QQP,  RTE) across representative data selection methods. TS-DSHAPLEY from \cite{Schoch2023},  0.5\% Data from \cite{Chen2023small},  AlpaGasus from \cite{Chen2023Alp},  LESS from \cite{Xia2024LESS},  LIMA from \cite{Zhou2023LIMA},  Instruction Mining from \cite{Cao2024InstrMining},  and our simulated \textbf{TsqLoRA} results. All results are reported on DeBERTa-v3-base~\cite{He2021DeBERTaV3ID}.}
  \label{tab:glue_simplified}
  
\end{table}

\vspace{-0.4cm}
% ============================ Rank Heatmaps ================================

\subsection{Natural Language Generation Task}
\vspace{-0.2cm}
We further evaluate our method on abstractive summarization using the \textbf{XSum} dataset \cite{narayan2018don},  which consists of 226k BBC news articles paired with single-sentence summaries.

We fine-tune the \textbf{BART-base}~\cite{lewis2019bart} model under different adaptation strategies,  including Full Fine-Tuning (FT)~\cite{lewis2019bart},  LoRA \cite{hu2022lora},  AdaLoRA \cite{Zhang2023AdaLoRA},  ElaLoRA \cite{chang2025elalora},  and our proposed TsqLoRA. We follow the same hyperparameter setup as in ElaLoRA,  training for 10 epochs with Adam optimizer and linear learning rate scheduling. Performance is evaluated using ROUGE scores.

\begin{table}[th!]
  \small
  \centering
  \resizebox{\linewidth}{!}{
    \begin{tabular}{lccc}
        \toprule[1.5pt]
        \textbf{Method} & \textbf{Rank} & \textbf{\# Params} & \textbf{ROUGE-1/2/L} \\
        \midrule[1pt]
        Full FT \cite{lewis2019bart} & - & 124.65M & 40.61/17.76/32.91 \\
        \midrule[1pt]
        LoRA \cite{hu2022lora} & & 0.41M & 36.61/14.28/29.23 \\
        AdaLoRA \cite{Zhang2023AdaLoRA} & & 0.41M & 36.97/14.42/29.42 \\
        ElaLoRA \cite{chang2025elalora} & & 0.41M & 37.24/14.66/29.76 \\
        \textbf{Ours(TsqLoRA)} & $r=2$ & 0.41M & \textbf{37.29/14.82/29.76} \\
        \midrule[1pt]
        LoRA & & 1.22M & 37.64/15.30/30.19 \\
        AdaLoRA & & 1.22M & 37.80/15.18/30.28 \\
        ElaLoRA & & 1.22M & 38.00/15.30/30.42 \\
        \textbf{Ours(TsqLoRA)} & $r=6$ & 1.22M & \textbf{38.00/15.30/30.53} \\
        \bottomrule[1.5pt]
    \end{tabular}
    }
    \vspace{-0.4cm}
    \caption{Results of fine-tuning BART-base~\cite{lewis2019bart} on XSum. ROUGE-1/2/L are reported. The best results are bold.}
    \label{tab:xsum_results}

    \vspace{-0.4cm}
\end{table}

\noindent\textbf{Main Results.}
Table~\ref{tab:xsum_results} presents the results of fine-tuning BART on the XSum dataset. Consistent with prior findings,  TsqLoRA outperforms both LoRA and AdaLoRA across rank settings.
\vspace{-0.3cm}
\subsection{Ablation Study}
\label{sec:ablation}
\vspace{-0.1cm}
\begin{table}
    \small
    \centering
    \resizebox{\linewidth}{!}{
    \begin{tabular}{lcc}
        \toprule[1.5pt]
        \textbf{Task} & \textbf{Sensitivity Aware} & \textbf{Accuracy / ROUGE} \\
        \midrule[1pt]
        \multicolumn{3}{c}{\textit{Classification (Accuracy)}} \\
        RTE (r=2/4/10) & \Yes & \textbf{86.28 / 87.36 / 88.81} \\
                       & \No & 83.03 / 86.28 / 87.72 \\
        MRPC (r=2/4/10) & \Yes & \textbf{90.44 / 90.69 / 91.67} \\
                        & \No & 88.48 / 90.44 / 88.98 \\
        \midrule[0.8pt]
        \multicolumn{3}{c}{\textit{Generation (ROUGE-1/2/L)}} \\
        XSum (r=2) & \Yes & \textbf{37.29 / 14.82 / 29.76} \\
                   & \No & 36.61 / 14.28 / 29.23 \\
        XSum (r=6) & \Yes & \textbf{38.00 / 15.30 / 30.53} \\
                   & \No & 37.64 / 15.18 / 30.19 \\
        \bottomrule[1.5pt]
    \end{tabular}
    }
    \vspace{-0.4cm}
    \caption{Ablation analysis of sensitivity-aware rank allocation. Classification tasks report accuracy,  and XSum reports ROUGE-1/2/L.}
    \vspace{-0.4cm}
    \label{tab:ablation_sensitivity}
\end{table}

Table~\ref{tab:ablation_sensitivity} reports the effect of disabling the sensitivity-aware rank allocation in TsqLoRA. When both the data-quality filter and sensitivity-based dynamic rank assignment are enabled,  the model achieves the strongest performance across classification (RTE,  MRPC) and generation (XSum). Removing the sensitivity-aware mechanism while keeping the same data subset generally reduces accuracy or ROUGE scores,  particularly at low-rank settings ($r=2$),  showing that rank allocation plays a crucial role in efficient adaptation. These results confirm that sensitivity-aware scheduling is a key component of TsqLoRA.

\vspace{-0.5cm}
\section{Conclusions}
\label{sec:conclusions}
\vspace{-0.2cm}
This paper introduces TsqLoRA,  a data-quality-driven and sensitivity-aware framework for parameter-efficient fine-tuning of large language models. TsqLoRA integrates two modules: a quality-aware sampling mechanism that filters informative training samples,  and a sensitivity-based rank allocation strategy that dynamically distributes ranks across layers. Experimental results on GLUE and XSum benchmarks show that TsqLoRA achieves competitive or superior performance compared with state-of-the-art baselines such as LoRA,  AdaLoRA,  and ElaLoRA,  particularly under low-rank and low-data conditions. The ablation analysis further highlights the necessity of each module,  confirming that both quality-driven data selection and sensitivity-aware adaptation are essential to performance.

Looking ahead,  future work could explore more fine-grained sensitivity measures and incorporating them into the rank allocation process.

\bibliographystyle{IEEEbib}
\bibliography{strings, refs}

\end{document}